\pdfoutput=1

\documentclass[11pt]{article}

\usepackage[final]{acl}

\usepackage{times}
\usepackage{latexsym}
\usepackage{booktabs}
\usepackage{tcolorbox}

\usepackage[T1]{fontenc}

\usepackage[utf8]{inputenc}

\usepackage{microtype}

\usepackage{inconsolata}

\usepackage{graphicx}
\usepackage{multirow}
\usepackage{longtable}
\usepackage{xspace}
\usepackage{tablefootnote}
\usepackage{pifont}
\usepackage{makecell}

\title{
\textsc{HumVI}\xspace: A Multilingual Dataset for Detecting Violent Incidents Impacting Humanitarian Aid}

\newcommand{\core}{core\xspace}
\newcommand{\expansion}{expansion\xspace}
\newcommand{\insecurityinsight}{Insecurity Insight\xspace}

\author{Hemank Lamba$^1$, Anton Abilov$^1$, Ke Zhang$^1$, Elizabeth M. Olson$^1$, \\
\textbf{Henry K. Dambanemuya$^3$ \thanks{Work done during author's internship at Dataminr.}, João C. Bárcia$^2$, David S. Batista$^2$,} \\ \textbf{Christina Wille$^2$, Aoife Cahill$^1$, Joel Tetreault$^1$, Alex Jaimes$^1$}\\
$^1$ Dataminr, Inc., $^2$ Insecurity Insight,  $^3$ Northwestern University\\
\{hlamba,aabilov,kzhang,elizabeth.olson,acahill,jtetreault,ajaimes\}@dataminr.com\\hdambane@u.northwestern.edu, christina.wille@insecurityinsight.org \\ \{joaobarcia, dsbatista\}@gmail.com
}

\begin{document}
\maketitle
\begin{abstract}

Humanitarian organizations can enhance their effectiveness by analyzing data to discover trends, gather aggregated insights, manage their security risks, 
support decision-making, and inform advocacy and funding proposals. However, data about violent incidents with direct
impact and relevance for humanitarian aid operations is not readily available. %
An automatic data collection and NLP-backed classification framework 
aligned with humanitarian perspectives can help bridge this gap. %
In this paper, we present \textsc{HumVI}\xspace ~-- a dataset comprising news articles in three languages (English, French, Arabic) containing instances of different types of violent incidents categorized by the humanitarian sector they impact, e.g., aid security, education, food security, health, and protection. Reliable labels were obtained for the dataset by partnering with a data-backed humanitarian organization, \insecurityinsight. 
We provide multiple benchmarks for the dataset, employing various deep learning architectures and techniques, including data augmentation and mask loss, to address different task-related challenges, e.g., domain expansion.
The dataset is publicly available at \href{https://github.com/dataminr-ai/humvi-dataset}{https://github.com/dataminr-ai/humvi-dataset}.%

\end{abstract}

\section{Introduction}
\label{sec:introduction}
Violent events that impact humanitarian efforts, such as the looting of aid trucks and the kidnapping of aid workers, frequently hinder the delivery of life-saving aid and protection efforts, with devastating consequences for conflict-affected populations. Conflicts also severely disrupt existing food systems, healthcare, and education structures ~\cite{gates2012development}, leading to food insecurity, malnutrition ~\cite{martin2019food}, increased mortality due to inadequate healthcare ~\cite{garry2020armed}, and significant educational disruptions for children ~\cite{kadir2019effects}. The systematic collection and collation of information about such events from multilingual data sources are not just crucial, but urgent, to ensure that decision-makers have access to the right information to support humanitarian operations and adequately respond to local needs.

\begin{figure}
\includegraphics[width=1.0\linewidth]{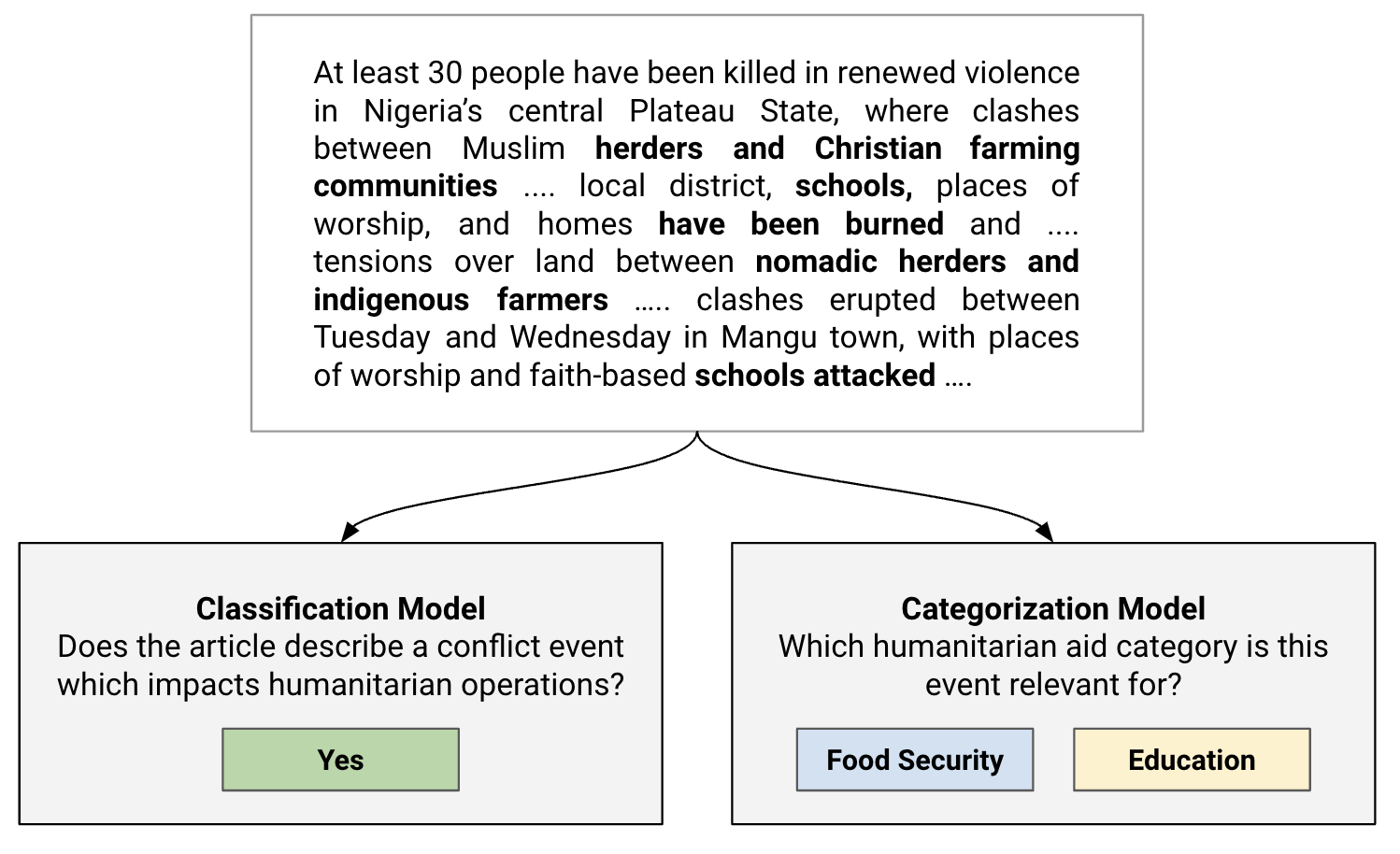}
\caption{A sample input/output to the two NLP models for (i) relevance classification, followed by (ii) categorization of the event into humanitarian aid categories.}
\label{fig:example}
\end{figure}

Aggregated information about 
violent events that impact humanitarian efforts 
can be operationalized by humanitarian organizations to systematically inform a wide range of different responses, including security risk management, program planning, and advocacy prioritization. Resources are severely limited during crisis response, and manual data collection in conflict-affected and highly insecure environments is rarely a priority. The use of NLP techniques can readily reduce the labor and resources required to maintain such a data collection pipeline. Importantly, it can also be carried out remotely, thus reducing the security exposure of those supporting a humanitarian response through data collection and analysis. Most NLP capacity in the humanitarian context has focused on English, limiting the use of models in conflict-affected countries where information is shared in French, Arabic, and other languages.

In order to process large amounts of disparate data to detect violent events, it is essential to build NLP models that can automatically detect such incidents.
Next, these incidents should be
tagged appropriately with the relevant type of humanitarian operation, e.g., an event where aid workers are threatened or harmed would be tagged as \textit{aid security}.

We identify two primary gaps in the current publicly available datasets:
(1) Most are focused on identifying a single type of event, such as disaster-related information~\cite{alam2021crisisbench,alam2018crisismmd}, specific threats against human rights defenders~\cite{ran2023new}, or civil unrest~\cite{delucia2023multi}. Even when datasets cover a wide range of events and associated tags~\cite{clionadh2010acled}, these tags often do not indicate the specific humanitarian sectors impacted by the event and lack information related to downstream humanitarian efforts, such as \textit{healthcare} or \textit{food security}~\cite{trivedi2020eco}. %
(2) Most of the humanitarian event detection datasets are disproportionately English and this has been identified as a key issue for improving adoption of NLP techniques in the humanitarian sector~\cite{kreutzer2019improving,rocca2023natural}. The dataset presented in this paper is multilingual, and we hope it will
support and encourage
NLP practitioners to create multilingual models for humanitarian purposes.

To address these two issues, 
we introduce a new dataset \textsc{HumVI}\xspace (HUManitarian Violent Incidents). This dataset comprises $17,497$ articles in three languages, with each article tagged for its relevance and, if relevant, categorized by the key humanitarian response sector where it is relevant, %
i.e., aid security, education, food security, health, and protection (a sample is shown in Figure~\ref{fig:example}). %

There are three qualities of this dataset that ensure its value to the field.

\noindent
\textbf{Humanitarian Expert Verified}.
A key challenge of building such a dataset is obtaining labels from the end-users, i.e. humanitarian experts, who will consume the tagged dataset to provide aid-centred event monitoring and context analysis. To solve this, we partner with \insecurityinsight\footnote{\href{https://insecurityinsight.org/}{https://insecurityinsight.org/}}, a data-driven humanitarian to humanitarian (h2h) organization that has expertise in applying and consuming tagged news articles in order to support the aid sector with aid-focused information.

\noindent
\textbf{AI for Social Good}. Recently, researchers have focused on multiple problems under the umbrella of AI for Social Good~(AI4SG)~\cite{shi2020artificial}. However, some of the SDGs~(Sustainable Development Goals) are under-represented within AI community efforts~\cite{gonzalez2023beyond}, including SDG 2: \textit{Zero Hunger}~\cite{fund2015sustainable}. A focus of our work is \textit{food security}, which aims to tag conflict events impacting food security. We believe that having a labeled dataset directly related to this SDG will 
encourage more NLP researchers to develop technology to benefit organizations that rely on understanding the impact of this SDG.

\noindent
\textbf{Domain Expansion}. The dataset-building task was carried out in conjunction with \insecurityinsight. While the organization had previously developed tagging systems for health, education, protection, and aid security events, expanding to a new category~(food security) and new languages~(French and Arabic) was an obstacle for them, given their limited staffing. This mirrors a 
challenging problem in the NLP space: domain expansion, \textit{i.e.}, expanding a machine learning model to work with new classes and on new input sources. In this paper, we provide two versions of the dataset: (a)~\textsc{HumVI}\xspace-\core and (b)~\textsc{HumVI}\xspace-\expansion. %
The core dataset focuses on four possible event types and is available only in English. However, the \expansion dataset extends the \core dataset by adding two new languages (French and Arabic) and one new category, food security. We believe that the \expansion dataset will enable NLP researchers to benchmark their solutions for the real-world challenges frequently encountered by humanitarian organizations while working in a capacity-constrained setting.

\noindent
To summarize, our contributions are follows:

\noindent
\textbf{[C1] Humanitarian Violent Event Dataset}. We provide a comprehensive multilingual dataset of violent events, labeled by humanitarian experts for relevancy to multiple key humanitarian aid sectors.

\noindent
\textbf{[C2] AI for Good}. To the best of our knowledge, this is the only dataset that provides instances of conflict events that might lead to food insecurity, 
an under-represented research area. 

\noindent
\textbf{[C3] Domain Expansion}. \textsc{HumVI}\xspace -\expansion provides a dataset that is representative of a challenging real world problem, \textit{i.e.}, resource-constrained domain expansion~\cite{yang2022learning}.

\noindent
\textbf{[C4] Baseline Experiments}. We show leading NLP models perform on this dataset, thus setting a baseline for future work while showcasing the complexities and challenges of adapting NLP techniques to real world scenarios.

\noindent
We make the dataset and associated repository public at \href{https://github.com/dataminr-ai/humvi-dataset}{https://github.com/dataminr-ai/humvi-dataset}.

\section{Related Work}

\subsection{Event Detection}
The development of large-scale geographically and temporally disaggregated conflict-event datasets emerged in the early 2010s, in large part to facilitate research on conflict dynamics. 
Among the earliest and best known are the Armed Conflict Location and Event Dataset (ACLED)~\cite{clionadh2010acled}, Social Conflict Analysis Database (SCAD)~\cite{salehyan2012social}, the Uppsala Conflict Data Program (UCDP)~\cite{sundberg2013ucdp}, and GDELT (Global Database of Events, Language and Tone)~\cite{leetaru2013gdelt}. These datasets are widely leveraged by humanitarian practitioners for conflict research, early warning, and crisis response~\cite{donnay2019integrating, sur_evaluation_2019, hegre2019views, penson2024data, ocha2024multihazard}. 
Most existing datasets are generic and extensive, covering events from various geographies. In contrast, \textsc{HumVI}\xspace focuses on events with direct humanitarian impact and relevance to humanitarian aid operations. The applied tags indicate the specific humanitarian aid sector(s) where the event is relevant.%

Additionally, another category of event detection dataset exists that is more specific and task-oriented. Recent work include datasets for detecting attacks against human right defenders~\cite{ran2023new}, monitoring infrastructure construction that threatens environmental conservation~\cite{keh2023newspanda}, detecting gun violence related attacks~\cite{pavlick2016gun}, classifying types of crime in social media messages~\cite{JarquinVasquezFAEPMS23}, detecting human rights violations~\cite{pilankar2022detecting}, and tracking migrant deaths and disappearances~\cite{brian2014migrant}. Though these datasets are carefully curated, they are limited to detecting events that are related to a single specific event type. 
Our dataset, on the other hand, provides relevance labels for several types of violent events that are key for contextualizing and informing the timely delivery of humanitarian aid in conflict settings.

The real-time nature of social media has made it a popular data source for detecting events in real time~\cite{fedoryszak2019real}. Most of the work has been centered on detecting crisis events with the goal of informing appropriate humanitarian response. \citet{alam2021crisisbench} proposed \emph{CrisisBench}, a dataset comprising 310K tweets annotated for informativeness and humanitarian response for multiple different types of crisis events, e.g., flood, earthquake, etc. Similarly,~\citet{parraga2021urbangency} created a dataset comprising 25K Spanish tweets indicating emergency or non-emergency related content in Ecuador~\cite{parraga2021urbangency}. Beyond classification, multiple new tasks like summarizing the crisis related tweets for effective humanitarian response have also been proposed~\cite{yela2021multihumes, faghihi2022crisisltlsum}. Though useful, event detection on social media platforms often face challenges associated with veracity, (\textit{i.e.}, able to filter out potentially unverified information or disinformation) and velocity, (\textit{i.e.}, the ability to handle rapid streams of messages at scale)~\cite{panagiotou2016detecting}, which are very different from the challenges associated with our dataset.%

Our work is closest to HumSet~\cite{fekih2022humset} which is a multilingual dataset of humanitarian response documents where each entry has been annotated for different sectors, and its impact and needs. HumSet is associated with humanitarian response documents which are typically reports generated by actors in the humanitarian sector (NGOS, UN agencies, etc), whereas \textsc{HumVI}\xspace is focused on news articles and can be used to inform the humanitarian response reports. We summarize major differences in Table~\ref{table:related_work}.

\begin{table}[]
\small{\begin{tabular}{@{}l|cccc@{}}
\toprule
\multirow{2}{*}{Datasets} & \multicolumn{4}{c}{Properties}             \\ \cmidrule{2-5}
                          & \multicolumn{1}{c}{\begin{tabular}[c]{@{}c@{}}News\\ Domain\end{tabular}} & \multicolumn{1}{c}{\begin{tabular}[c]{@{}c@{}}Usecase\\ Tags\end{tabular}} & \multicolumn{1}{c}{\begin{tabular}[c]{@{}c@{}}Multi\\ Lingual\end{tabular}} & \multicolumn{1}{c}{\begin{tabular}[c]{@{}c@{}}Manual\\ Labels\end{tabular}} \\ \midrule
GDELT  & \textcolor{green}{\ding{52}} & \textcolor{red}{\ding{56}}& 
         \textcolor{green}{\ding{52}} & \textcolor{red}{\ding{56}} \\
ACLED & \textcolor{green}{\ding{52}} & \textcolor{red}{\ding{56}} &
        \textcolor{green}{\ding{52}} & \textcolor{green}{\ding{52}} \\
CrisisBench & \textcolor{red}{\ding{56}} & \textcolor{red}{\ding{56}}         & \textcolor{red}{\ding{56}} & \textcolor{green}{\ding{52}} \\
HumSet & \textcolor{red}{\ding{56}} & \textcolor{green}{\ding{52}} &
        \textcolor{green}{\ding{52}} & \textcolor{green}{\ding{52}} \\
\midrule
\textbf{\textsc{HumVI}\xspace} & \textbf{\textcolor{green}{\ding{52}}} & \textbf{\textcolor{green}{\ding{52}}} & \textbf{\textcolor{green}{\ding{52}}} & \textbf{\textcolor{green}{\ding{52}}} \\ 
\bottomrule
\end{tabular}
}
\caption{A brief overview of the datasets in this space in comparison to the proposed dataset.
}
\label{table:related_work}
\end{table}

\subsection{NLP for Social Good}
Advancements in NLP technologies have broadened their application scope across education~\cite{kasneci2023chatgpt,madnani2018automated}, assistants~\cite{de2020intelligent}, legal fields~\cite{martinez2023survey}, and more. With these advancements, researchers are increasingly prioritizing NLP applications for societal benefit~\cite{cowls2021definition}. Significant positive impacts have been achieved in healthcare~\cite{ghassemi2020review}, disaster response~\cite{arachie2020unsupervised,liu2021crisisbert,alam2021crisisbench}, climate change~\cite{luo2020detecting,diggelmann2020climate}, and mental health~\cite{perez2019makes,evensen2019happiness}, aligned with UN SDGs~\cite{gonzalez2023beyond}. Despite being rated as important~\cite{yang2020prioritizing}, %
some SDGs like SDG1 (No Poverty), SDG2 (Zero Hunger), SDG6 (Clean Water), and SDG13 (Climate Action) receive little attention from NLP researchers. Our dataset includes labeled instances relevant to humanitarian concerns, addressing issues such as food insecurity linked directly to SDG2 (Zero Hunger). This dataset can support NLP researchers exploring these critical areas.

\subsection{Domain Expansion}

A key challenge in NLP research is to develop multilingual models. 
Early methods included cross-lingual transfer of monolingual embeddings, aligned using unsupervised techniques~\cite{lample2017unsupervised, conneau2017word}. A natural extension was to create multilingual embeddings that can be used directly~\cite{ruder2019survey}. Following the success of transformer models, pre-trained multilingual models were created~\cite{devlin2018bert,ebrahimi2021adapt}. With the availability of large-scale data and compute, massive encoder-decoder models~\cite{liu2020multilingual, xue2020mt5} and then eventually decoder-only models~\cite{achiam2023gpt} were trained. These models can be leveraged in zero-shot or few-shot settings and still be performant without requiring any specific fine-tuning. Another set of approaches includes augmenting the training dataset by translating it into the target language
~\cite{edunov-etal-2018-understanding}.%

Another key challenge is to how to extend the model to new categories, where getting high-quality labeled data for training is expensive or unavailable. Data augmentation is one of the most popular techniques that transforms existing data in a reliable class-preserving manner to create new data~\cite{chen2023empirical}. Self-training~\cite{du2020self,scudder1965adaptive} is a promising technique that enables leveraging unlabeled datasets to create pseudo-labels which can be used to further improve the model. %
Other techniques like multi-task learning and consistency regularization have also been used to tackle this issue. However, more recently, the most popular methods have been to scale up language models and leverage their zero-shot or few-shot capabilities to achieve strong performance~\cite{brown2020language}. The dataset presented in this paper poses multilinguality and limited data learning as two of its main challenges. As part of our process of benchmarking the dataset, we apply the methods mentioned above and demonstrate their performance.

\section{Data Collection}
\subsection{Data Collection Overview}
\label{sec:data}

\begin{figure}[!tbp]
    \centering
    \includegraphics[width=\linewidth]{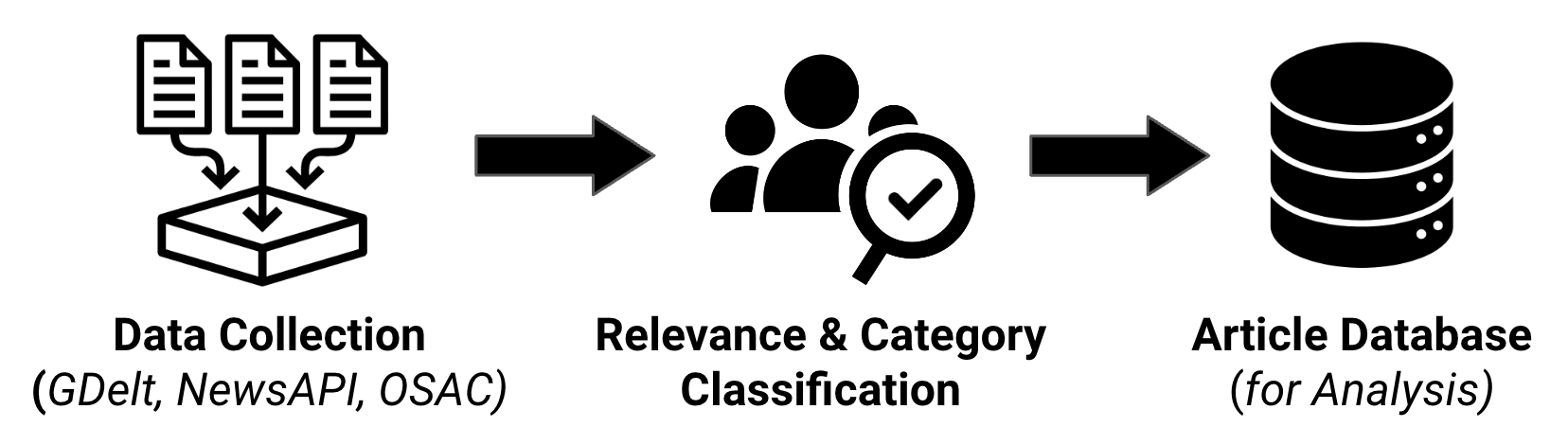}
    \caption{A schematic representation of \insecurityinsight's data processing system.}
    \label{fig:ii_architecture}
\end{figure}

\begin{table*}[!hbtp]
\small{
\begin{tabular}{lp{6.1cm}p{6.1cm}}
\toprule
Category & Description & Example (Title with URL in bracket) \\
\midrule
Aid Security \textit{(aid)} & Harms or threats towards aid agencies or aid workers. & Aid group says Israeli strike kills 7 of its workers in Gaza, including foreigners. [\href{https://tinyurl.com/4szy5d6d}{https://tinyurl.com/4szy5d6d}]
\\

Education \textit{(edu)} & Harms or threats towards education providers, education workers, education related infrastructure and students. 
& Nigeria: Gunmen Kidnap Nine Students in Delta. [\href{https://tinyurl.com/vduwuxyd}{https://tinyurl.com/vduwuxyd}]
\\

Food Security \textit{(food)}
& Harm or threats towards entities involved in food production, processing, and distribution, as well as damage to transport and energy infrastructure that supports the food supply chain.
& Benue buries 17 persons killed by `herders' as governor suggests solution. [\href{https://tinyurl.com/2jvjyxva}{https://tinyurl.com/2jvjyxva}]
\\

Health \textit{(hlth)} &  Harm or threats towards health providers or health workers. & FCT NMA raises concern over kidnapping of members. [\href{https://tinyurl.com/4uv49u7f}{https://tinyurl.com/4uv49u7f}]
\\

Protection \textit{(prtc)} & Harm or threats towards internally displaced persons and refugees. & Israeli Airstrike Kills 36 Palestinians During Suhoor Meal in Nuseirat Refugee Camp. [\href{https://tinyurl.com/5abcbaz7}{https://tinyurl.com/5abcbaz7}]
\\           

\bottomrule
\end{tabular}
}
\caption{Description of different categories along with examples. The shorthand used for the categories is in brackets. In particular, Food Security is the new category being added as part of \expansion dataset}
\label{table:labels}
\end{table*}

As mentioned in Section~\ref{sec:introduction}, we worked closely with \insecurityinsight to create \textsc{HumVI}\xspace. %
\insecurityinsight already had an established workflow for sourcing relevant articles, after which humanitarian experts classify them for relevance, and then tag them with the appropriate categories that capture the downstream humanitarian concern~(Figure~\ref{fig:ii_architecture}). %
They were also leveraging multiple ways to identify links to news articles, which we list below:

\noindent
\textbf{NewsAPI}. 
Researchers have extensively used NewsAPI~\cite{lisivick2018newsapi} to collect news articles for various purposes~\cite{keh2023newspanda,jain2024really}.  Similarly, for this dataset, NewsAPI was used to search for English articles about violent conflict incidents using a curated list of domain-specific keywords, detailed in Appendix \ref{sec:news_api}.

\noindent
\textbf{OSAC}. The Overseas Security Advisory Council\footnote{\href{https://www.osac.gov/Content/Browse/News}{https://www.osac.gov/Content/Browse/News}} is a US Department of State organization whose primary purpose is to share critical information related to security. As part of this effort, at regular intervals, they share English articles selected by OSAC analysts that are highly relevant to global security. This list is collected by \insecurityinsight, and added to the queue to be reviewed by their experts.%

\noindent
\textbf{Manual Collection}. Besides classifying articles as relevant and tagging them with appropriate categories, experts at \insecurityinsight also manually upload news articles to the database. These articles may have been missed by automatic scrapers, and hence need to be added to ensure more comprehensive coverage. %

\subsection{Source Expansion}
One of the key challenges that 
motivated the collection of \textsc{HumVI}\xspace 
was the need to expand the current data collection and tagging process to include a new category (food security) and new languages (French and Arabic), which were rarely covered in the original sources. To enhance sourcing, we added GDELT~\cite{leetaru2013gdelt}, a large open-source database of multilingual news article links. We query the GDELT Event Database for articles which are regionally tied to either Burkina Faso, Cameroon, the Central African Republic, the Democratic Republic of the Congo, Palestine, Haiti, Mali, Niger, Nigeria, Somalia, Syria or Yemen. We scrape the full article content and detect the article language, and English, French and Arabic articles are added to the database.

\subsection{Labeling Process}
Once the potentially relevant news articles are identified, the title and full text of the news article is collected. The human expert first annotates the relevance of the article (a binary classification). The relevance is determined by assessing whether the article describes a conflict event and the event belongs to one or more of the pre-defined humanitarian categories (aid security, education, food security, health and protection). %
For a category to be applied the article should describe reports of conflict events happening at a particular time and location, typically described as being carried out by a \textit{perpetrator} against a given \textit{person }\textit{who fulfills a specific humanitarian function }(\textit{e.g. }an aid or health worker) or \textit{essential} \textit{civilian infrastructure }(\textit{e.g. }aid convoys, hospitals, bakeries or schools) \footnote{Detailed definitions are provided here: \href{https://insecurityinsight.org/methodology-and-definitions}{https://insecurityinsight.org/methodology-and-definitions}}.
A brief overview of the categories is given in Table~\ref{table:labels} (detailed description in Appendix~\ref{sec:appendix_annotation}).

We obtained the labels in two ways: (1) On-the-job labeling (OTJ) and (2) Offline labeling (OFL). Data collected via OTJ originates from the partnering organization’s established production workflows and was collected live from the English-only data sources News API and OSAC. The scraped article title and content is reviewed by the expert annotators in their internal annotation tool before they determine whether the article is relevant and assign the event categories. 
For OFL, we worked with 7 humanitarian experts from \insecurityinsight to expand the data collection to include articles from GDELT in English, French, and Arabic. We follow similar annotation guidelines as in OTJ. The annotators performed the offline task in spreadsheets. They reviewed the scraped article title and full text before assigning one or more event categories as per the guidance. If none of the event categories are assigned, the sample is marked as non-relevant. 
\textsc{HumVI}\xspace contains OTJ labels for most training data and OFL labels for the test data.

\subsubsection{Quality Control}
\label{sec:quality-control}
Both methods of soliciting labels are carried out by a team of domain experts trained by \insecurityinsight. For OTJ labeling, the labeling %
correctness is ensured by significant training efforts, well defined guidelines, and the periodic review and correction by a supervising expert.  
OFL labeling was carried out by experts who have been trained similarly and were asked to follow similar guidelines.

\begin{table}[ht]
\centering
\small{
\begin{tabular}{l|ccc}
\toprule
\multirow{2}{*}{Label} & \multicolumn{3}{c}{Language}  \\ 
    & \begin{tabular}[c]{@{}c@{}}English\\ A=3, N=100\end{tabular} 
    & \begin{tabular}[c]{@{}c@{}}French\\ A=7 N=75\end{tabular}
    & \begin{tabular}[c]{@{}c@{}}Arabic\\ A=7 N=75\end{tabular}     \\ 
\midrule
Relevance   & 0.89 & 0.77 & 0.86 \\
\midrule
Aid         & 0.91 & 0.89 & 0.66 \\
Education   & 0.95 & 0.67 & 0.64 \\
Health      & 0.88 & 0.80 & 0.75 \\
Protection  & 0.96 & 0.78 & 0.88 \\
Food        & 0.81 & 0.59 & 0.69 \\
\bottomrule
\end{tabular}
}
\caption{Average pair-wise Kappa score for $N$ Samples in each language across $A$ annotators.}
\label{table:agreement}
\end{table}

\begin{table}[!hbtp]
\centering
\small{
\begin{tabular}{l|lll}
\toprule
\multirow{2}{*}{Source} & \multicolumn{3}{c}{Language}  \\ 
    & English & French & Arabic     \\ 
\midrule
News API & 7,486 & 0 & 0 \\
OSAC     & 7,069 & 0 & 0 \\
GDELT$^{*}$    & 867 & 894 & 900 \\
Manual$^{*}$   & 209 & 35 & 37 \\
\midrule
\midrule
Total & 15,631 & 929 & 937  \\
Relevant & 3,049 & 233 & 358  \\
\midrule
\midrule
Unlabeled & 41,589 & 4,689 & 8,977 \\
\bottomrule
\end{tabular}
}
\caption{Source and Language Distribution. $^{*}$ indicates the labels for data source were collected in OFL manner.}
\label{table:sourcelangdistribution}
\end{table}

\begin{table*}
\centering
\small{
\begin{tabular}{l|llllll|llllll}
\toprule
& \multicolumn{6}{c|}{Train Dataset}
& \multicolumn{6}{c}{Test Dataset}     \\
\toprule
lang & aid & edu & prtc & hlth & food & total
   & aid & edu & prtc & hlth & food & total  \\
\toprule
en & \textbf{518} & \textbf{1,101} & \textbf{256} & \textbf{900} & 241 & 14,593
   & \textbf{124} & \textbf{70}    & \textbf{77}  & \textbf{149} & 134 & 1,038     \\
fr & 14  & 1     & 7   & 9   & 44  & 133
   & 58  & 60    & 12  & 69  & 52  & 796       \\
ar & 10  & 2     & 22  & 20  & 47  & 137 
   & 90  & 57    & 91  & 106 & 72  & 800 \\

\bottomrule
\end{tabular}
}
\caption{Language and class distribution across train test split of the dataset. The table represents \textsc{HumVI}\xspace - \expansion, the \textbf{bolded} entries are a part of \textsc{HumVI}\xspace - \core.}
\label{table:dataset_splits}
\end{table*}

We also measured inter-annotator agreement rates in English, French and Arabic. As shown in Table~\ref{table:agreement}, the average annotator agreement rate is high (indicating substantial to near perfect agreement~\cite{landis1977measurement}), and only slightly worse for some annotators on the new category being introduced (food security). Extended annotation guidelines and details are provided in Appendix~\ref{sec:appendix_annotation}.

 We removed articles from the collected data if the crawl failed to retrieve the text or if the article contained no text. Additionally, we eliminated duplicate articles with identical lowercase text or identical URLs.

\subsection{Data Description}

\textsc{HumVI}\xspace includes $17,497$ labeled articles in English, French, and Arabic. 
The high-level statistics are listed in Table~\ref{table:sourcelangdistribution}. 
English articles are predominantly sourced from NewsAPI and OSAC, whereas French and Arabic articles were either sourced from GDELT or collected manually as shown in Table~\ref{table:sourcelangdistribution}. Note that the NewsAPI, OSAC and English articles are disproportionately high in our dataset because the majority of those instances were part of the standard workflow at \insecurityinsight 
when data collection began.
Labels for NewsAPI and OSAC articles have been obtained through OTJ labeling, whereas all GDELT and manually collected articles are labelled through OFL.

\section{Dataset Split and Variants}
\label{sec:data_splits_variants}
We provide two different variants of the dataset described in the previous section.

\noindent
\textbf{\textsc{HumVI}\xspace-\core}. The \core dataset is English only, and contains labels for four categories relevant to humanitarian work ~-- it does not include any label for food security. The dataset consists of $15,631$ news articles.

\noindent
\textbf{\textsc{HumVI}\xspace-\expansion}. The \expansion dataset is centered on the real world challenge of expanding to a new category (food security) and new languages (French, Arabic). This dataset variant is a superset of \textsc{HumVI}\xspace-\core, and contains $17,497$ articles. This dataset poses multiple interesting challenges for an NLP practitioner - (i) how to expand to new modeling classes and languages (ii) how to build performant models in a resource-constrained manner.

For both variants of the dataset, we temporally 
split the data on 2024-02-08 such that all of the articles before this date are part of the train set. The test set is collected exclusively through the OFL labeling process and consists of news articles collected after 2024-03-15. All news articles collected between 2024-02-08 (train cutoff) and 2024-03-15 (test start date) are part of an unlabeled dataset and are also released along with the paper.\footnote{The unlabeled dataset only contains articles from GDELT. We have found GDELT to be a superset of query based NewsAPI access and OSAC.} %
We present detailed numbers on the dataset split in Table~\ref{table:dataset_splits}. Note that in the training dataset there are very few instances of French and Arabic, however, in the test dataset, there are a similar number of articles for all three languages. 
Similarly, the presence of the food security tag is under-represented in the training dataset compared to other categories, but equally represented in the test dataset.

Along with \textsc{HumVI}\xspace-\expansion, we also release a large collection of unlabeled data that can be used to improve the model's performance through active learning~\cite{settles2009active} or through semi-supervised learning~\cite{learning2006semi}.

\section{Models}
\label{sec:models}
We experiment with multiple different model types to establish benchmarks for two tasks on \textsc{HumVI}\xspace: (1) relevance classification (i.e., predict if the given news article is relevant or not); and (2) category classification (i.e., predict one or more categories where news article might be relevant).

\subsection{Base Architectures}

\noindent
\textbf{Fine-tuned Models}. In our experiments, we train separate models for each task. However, they could also be modeled jointly by training a single model to predict both relevance and category labels -  which we leave as an open research direction. We experiment with multiple different model architectures to benchmark their performance on \textsc{HumVI}\xspace.

We experiment with fine-tuning multiple different transformer models, specifically monolingual (English) and multilingual variants of \texttt{DistilBERT}~\cite{sanh2019distilbert}, \texttt{BERT}~\cite{devlin2018bert} and \texttt{RoBERTa}~\cite{liu2019roberta, conneau2019unsupervised}.

\noindent
\textbf{LLM}. We benchmark five zero-shot LLM models on the dataset: two open-source models, \texttt{DBRX}~\cite{databricks_dbrx_2023} and \texttt{LLaMa3-70b}~\cite{llama3modelcard}, and three proprietary models, \texttt{GPT-4}~\cite{achiam2023gpt}, \texttt{GPT-4o}~\cite{openai_gpt4o_2024}, and \texttt{Mistral Large}~\cite{mistral_large_2024}. All models use $temperature = 0$ and the same chain-of-thought-style prompt, derived from the annotation protocol with minimal tuning.\footnote{The full prompt is available \href{https://anonymous.4open.science/r/davinci-dataset-F014/llm_prompt/prompts.yaml}{here}.} The LLM experiments are set up as a single \textit{implicit }category classification task: if an article is labeled with a category, it is considered relevant.

\subsection{Domain Expansion}

To handle domain expansion we experiment with the following two approaches.%

\noindent
\textbf{Translation Augmentation}.
In order to adapt to new languages, we create synthetically labeled data by translating English samples into French and Arabic~\cite{edunov-etal-2018-understanding}.

\begin{table}[!hbtp]
\centering
\small{
\begin{tabular}{l|c|cccc}
\toprule
Model & relev. & aid & edu & hlth & prtc \\
\midrule
\texttt{DistilBERT} & 0.84 & 0.86 & 0.86 & 0.83 & 0.83 \\
\texttt{BERT} &   0.84 & \textbf{0.87} & 0.83 & 0.84 & \textbf{0.84}\\
\texttt{RoBERTa} & \textbf{0.86} & \textbf{0.87} & \textbf{0.87} & \textbf{0.85} & \textbf{0.84}   \\
\midrule
\texttt{DBRX} & 0.67 & 0.41 & 0.26 & 0.54 & 0.31   \\
\texttt{Llama-3-70B} & 0.72 & 0.60 & 0.66 & 0.70 & 0.37  \\
\texttt{Mistral-Large} & 0.79 & 0.75 & 0.69 & 0.76 & 0.42   \\
\texttt{GPT-4-Turbo} & 0.80 & 0.76 & 0.69 & 0.75 & 0.47 \\
\texttt{GPT-4-o} & 0.81 & 0.76 & 0.75 & 0.72 & 0.53 \\
\bottomrule
\end{tabular}
}
\caption{F1 Scores of various models on \textsc{HumVI}\xspace-\core for both relevance and category classification.}
\label{table:results_core}
\end{table}

\noindent
\textbf{Masking Loss}.
Since labeling for the food security tag started post the train cutoff date in OFL manner,
most instances in the \textsc{HumVI}\xspace -\expansion training dataset lack this label. To prevent learning from partially labeled data, we further use label loss masking~\cite{duarte2021plm} in category classification. The Binary Cross Entropy (BCE) loss is average pooled only over labeled categories, excluding food security.

\begin{table}[!hbtp]
\centering
\small{
\begin{tabular}{l|ccc}
\toprule
{Model} & English & French & Arabic   \\
\midrule
\texttt{DistilBERT} & 0.82 & 0.74 & 0.79 \\
\texttt{BERT} & \textbf{0.83} & 0.76 & 0.80 \\
\texttt{XLM-RoBERTa} & 0.82 & \textbf{0.77} & \textbf{0.83}    \\
\midrule
\texttt{DBRX} & 0.71 & 0.62 & 0.66  \\
\texttt{Llama-3-70B} & 0.73 & 0.60 & 0.67   \\
\texttt{Mistral-Large} & 0.78 & 0.76 & 0.74    \\
\texttt{GPT-4-Turbo} & 0.78 & 0.74 & 0.71 \\
\texttt{GPT-4-o} & 0.80 & 0.69 & 0.76  \\
\bottomrule
\end{tabular}
}
\caption{F1-Scores on \textsc{HumVI}\xspace-\expansion for relevance classification.}
\label{table:results_expansion_relevance}
\end{table}

\begin{table*}[!hbtp]
\centering
\small{
\begin{tabular}{l|ccccc|ccccc|ccccc}
\toprule
\multirow{2}{*}{Model} & \multicolumn{5}{c|}{English} & \multicolumn{5}{c|}{French} & \multicolumn{5}{c}{Arabic}   \\
& aid & edu & hlth & prtc & food & aid & edu & hlth & prtc & food & aid & edu & hlth & prtc & food   \\
\midrule
\begin{tabular}[c]{@{}c@{}}\texttt{Distil-}\\\texttt{BERT}\end{tabular}
        & 0.83 & 0.87 & 0.79 & 0.81 & 0.45
        & 0.79 & 0.87 & 0.80 & 0.67 & 0.34
        & 0.79 & 0.82 & 0.76 & 0.85 & 0.32  
                    \\

\texttt{BERT}       
    & \textbf{0.85} & \textbf{0.88} & 0.82 & 0.82 & 0.51
    & \textbf{0.84} & \textbf{0.88} & \textbf{0.85} & \textbf{0.73} & 0.44
    & 0.80 & \textbf{0.86} & 0.79 & \textbf{0.84} & 0.28 \\

\texttt{XLM-R}
    & \textbf{0.85} & \textbf{0.88} & \textbf{0.82} & \textbf{0.84} & \textbf{0.58}
    & \textbf{0.84} & \textbf{0.88} & 0.82 & 0.68 & \textbf{0.64}
    & \textbf{0.82} & \textbf{0.86} & \textbf{0.83} & \textbf{0.84} & \textbf{0.41}  \\ 
\midrule
\texttt{DBRX}       & 0.41 & 0.26 & 0.54 & 0.31 & 0.37
                    & 0.31 & 0.46 & 0.59 & 0.08 & 0.27 
                    & 0.27 & 0.24 & 0.51 & 0.35 & 0.20  \\
                    
\texttt{Llama3}& 0.60 & 0.66 & 0.70 & 0.37 & 0.41
                    & 0.53 & 0.76 & 0.69 & 0.09 & 0.33 
                    & 0.46 & 0.51 & 0.65 & 0.39 & 0.26  \\

\texttt{Mistral}& 0.75 & 0.69 & 0.76 & 0.42 & 0.44
                    & 0.66 & 0.80 & 0.78 & 0.19 & 0.41 
                    & 0.60 & 0.55 & 0.67 & 0.46 & 0.30  \\

\texttt{GPT-4T}& 0.76 & 0.69 & 0.75 & 0.47 & 0.29
                    & 0.62 & 0.80 & 0.77 & 0.24 & 0.35 
                    & 0.70 & 0.56 & 0.76 & 0.24 & 0.18  \\

\texttt{GPT-4o}    & 0.76 & 0.75 & 0.72 & 0.53 & 0.34
                    & 0.67 & 0.82 & 0.76 & 0.15 & 0.35 
                    & 0.70 & 0.66 & 0.69 & 0.42 & 0.24  \\
\bottomrule
\end{tabular}
}
\caption{F1-Scores on \textsc{HumVI}\xspace-\expansion for category classification.}
\label{table:results_expansion_categorization}
\end{table*}

\section{Results}
\noindent
\textbf{Metric of Choice}. 
While there is a range of metrics that could evaluate model performance, we 
settled on the following use-case-specific metrics together with \insecurityinsight.
For the relevance classifier, we compute precision at a minimum recall of 0.8 to surface more relevant content. For the category classifier, we optimize for precision and compute recall at a minimum precision of 0.8. We report the harmonic mean F1 score of these metrics averaged across five runs. 
For LLMs, we compute the F1 score directly based on the predictions without applying a threshold.

\noindent
\textbf{Results on \textsc{HumVI}\xspace-\core}. We report results in Table~\ref{table:results_core}. The fine-tuned transformers-based methods perform best for both relevance and categorization tasks. The large language models did not perform as well for either the relevance or category classification tasks. We believe that the relatively low performance of LLMs can be attributed to the prompt being too large and occupying a lot of context length. Future work could include reducing the prompt length, and then running few shot inference to try and improve performance. 

\begin{table}
\centering
\small{
\begin{tabular}{ccccc}
\toprule
Lang & Mode & \texttt{DistilBERT} & \texttt{BERT} & \texttt{XLM-RoBERTa}  \\
\midrule
\multirow{2}{*}{English} & Base & 0.81 & 0.82 & 0.83  \\
                         & wAug & 0.82 & 0.83 & 0.82 \\
\midrule
\multirow{2}{*}{French} & Base & 0.69 & 0.76 & 0.81    \\
                        & wAug   & 0.74 & 0.76 & 0.77    \\
\midrule
\multirow{2}{*}{Arabic} & Base & 0.69 & 0.75 & 0.81 \\
                        & wAug & 0.79 & 0.80 & 0.83  \\
\bottomrule
\end{tabular}
}
\caption{Relevance classification F1 scores of finetuned models on \textsc{HumVI}\xspace-\expansion before (\textbf{Base}) and after (\textbf{wAug}) translation augmentation.}
\label{table:ablation_augmentation}
\end{table}
\noindent
\textbf{Results on \textsc{HumVI}\xspace-\expansion}. We note that \texttt{XLM-RoBERTa} performs the best in adapting to the new category and new languages with limited labeled data for both relevance classification (Table~\ref{table:results_expansion_relevance}) and categorization (Table~\ref{table:results_expansion_categorization}). Unsurprisingly, all models struggle with performance on the new category (food security). The challenge of adapting to new languages (French, Arabic) is however manageable %
by leveraging multilingual pretrained transformers. LLMs do not perform as well as the finetuned models especially for the French and Arabic protection category. For LLMs, the proprietary models outperform the open-source ones, sometimes by a large margin.  For example, the difference in performance between \texttt{Llama-3} and \texttt{GPT-4o} is 1.2x for the English aid security category, and up to 1.5x for the Arabic aid security category. 

\noindent
\textit{Effect of Translation Augmentation}. We report the effect of translation augmentation on relevance classification performance for fine-tuned models for \textsc{HumVI}\xspace-\expansion in Table~\ref{table:ablation_augmentation}. 
The results for categorization are presented in Table~\ref{table:results_expansion_classification}. We note that, in general, translation augmentation improves model performance for new languages introduced, particularly Arabic. Similar trends are also observed for other fine-tuned transformer models for the category classification task.

\noindent
\textit{Effect of Masking Loss}. 
In general, we see that the application of masking loss improves the performance of the fine-tuned models (Table~\ref{table:results_expansion_classification}) in certain cases. However, for food security, the performance increases for English, but decreases considerably for French and Arabic.

For future modeling work, we report some misclassification themes (Appendix~\ref{sec:appendix_error})

\begin{table*}
\centering
\tabcolsep=0.12cm
\small{
\begin{tabular}{ll|lllll|lllll|lllll}
\toprule
& & \multicolumn{5}{c|}{English} & \multicolumn{5}{c|}{French} &
\multicolumn{5}{c}{Arabic}  \\
\midrule
Architecture & Mode 
& Aid & Edu & Hlth & Prtc & Food 
& Aid & Edu & Hlth & Prtc & Food 
& Aid & Edu & Hlth & Prtc & Food \\
\midrule
\multirow{4}{*}{\texttt{distilBERT}} &
Base & 0.82 & 0.87 & 0.81 & 0.83 & 0.28
     & 0.70 & 0.82 & 0.73 & 0.16 & 0.13
     & 0.64 & 0.70 & 0.69 & 0.16 & 0.13 \\
&wAug & 0.84 & 0.87 & 0.79 & 0.81 & 0.33 
     & 0.84 & 0.87 & 0.82 & 0.58 & 0.46
     & 0.78 & 0.82 & 0.76 & 0.85 & 0.26 \\
&wMask& 0.83 & 0.86 & 0.80 & 0.82 & 0.41
     & 0.76 & 0.83 & 0.76 & 0.15 & 0.05
     & 0.71 & 0.72 & 0.72 & 0.33 & 0.19 \\
&wBoth& 0.83 & 0.87 & 0.79 & 0.81 & 0.45
     & 0.79 & 0.87 & 0.80 & 0.67 & 0.34
     & 0.79 & 0.82 & 0.76 & 0.85 & 0.32  \\
\midrule
\multirow{4}{*}{\texttt{BERT}} &
Base & 0.85 & 0.88 & 0.82 & 0.82 & 0.41
     & 0.79 & 0.88 & 0.81 & 0.44 & 0.52
     & 0.77 & 0.81 & 0.78 & 0.83 & 0.31 \\
&wAug & 0.85 & 0.87 & 0.83 & 0.81 & 0.46
      & 0.85 & 0.88 & 0.85 & 0.40 & 0.52
      & 0.80 & 0.85 & 0.80 & 0.83 & 0.10 \\
&wMask & 0.85 & 0.88 & 0.83 & 0.83 & 0.53
       & 0.82 & 0.88 & 0.81 & 0.49 & 0.42
       & 0.77 & 0.81 & 0.79 & 0.84 & 0.25   \\
&wBoth & 0.85 & 0.88 & 0.82 & 0.82 & 0.51
       & 0.84 & 0.88 & 0.85 & 0.73 & 0.44
       & 0.80 & 0.86 & 0.79 & 0.84 & 0.28  \\
\midrule
\multirow{4}{*}{\texttt{XLM-RoBERTa}} &
Base & 0.85 & 0.87 & 0.83 & 0.83 & 0.48
     & 0.83 & 0.87 & 0.83 & 0.45 & 0.65
     & 0.81 & 0.83 & 0.82 & 0.72 & 0.52  \\
&wAug & 0.86 & 0.88 & 0.82 & 0.82 & 0.46
      & 0.84 & 0.88 & 0.83 & 0.68 & 0.65
      & 0.83 & 0.86 & 0.84 & 0.83 & 0.29    \\
& wMask & 0.86 & 0.87 & 0.82 & 0.83 & 0.52
       & 0.83 & 0.88 & 0.81 & 0.62 & 0.50
       & 0.80 & 0.83 & 0.82 & 0.80 & 0.45   \\
&wBoth & 0.85 & 0.88 & 0.82 & 0.84 & 0.58
       & 0.84 & 0.88 & 0.82 & 0.68 & 0.64
       & 0.82 & 0.86 & 0.83 & 0.84 & 0.41   \\
\bottomrule
\end{tabular}}
\caption{F1 Categorization Performance for \textsc{HumVI}\xspace-\expansion. Different variations of the fine-tuned models with Augmentation (wAug), Mask Loss (wMask) and both (wBoth) are shown.
}
\label{table:results_expansion_classification}
\end{table*}

\section{Conclusions}
\label{sec:conclusions}

\noindent

In this paper, we presented \textsc{HumVI}\xspace, a multilingual dataset of $17K$ news articles classified by their relevance to humanitarian aid efforts and tagged with appropriate categories. Developed in collaboration with \insecurityinsight, this dataset addresses real-world challenges in optimizing data workflows with limited resources using NLP techniques. We trained and evaluated multiple NLP models for relevance and event classification, showing that both fine-tuned models and zero-shot LLMs can perform well, leaving room for improvement. 

We believe that \textit{humanitarian organizations} can utilize this dataset to classify real-time news articles based on relevance and tag them with specific humanitarian aid sector relevant categories. The results can help determine whether LLMs or finetuned transformer-based models are suitable for their processes. Additionally, organizations needing extra tagging for humanitarian events associated with the text can benefit from the models trained using the provided data and code.

For \textit{NLP researchers} we hope that they will find this dataset a valuable resource for addressing real-world problems in the AI4SG space, particularly for developing techniques in resource-limited scenarios. The dataset presents interesting challenges, with significant potential for improving baseline performance.

\section{Limitations}
Like any data-based study, the dataset presented here is subject to multiple limitations. We take a critical approach for the dataset we have created and list the limitations and biases that could exist at each stage of the process followed. For \textit{data collection}, our dataset comes from a limited set of countries and hence might not be readily extensible to different parts of the world that we have not covered. Additionally, the differences in event reporting and content might need to be analyzed before being applied to a new geographic area. Similarly, we leverage three different ways of obtaining news article links. 
However, these may be systematically biased towards certain types of content, and they may ignore other types of content, e.g., hyperlocal news reporting in local languages that these link curation systems have not included. 
We rely heavily on leveraging GDELT to retrieve news articles in different languages from varied geographies. However, GDELT’s coverage might not be equitable across different languages and regions \cite{leetaru2013gdelt}, resulting in under-representation of certain geographies.
For \textit{annotation}, we leverage the labels that have been defined by \insecurityinsight. Though the labels were created with due deliberation and discussion with humanitarian experts, they are still tuned for \insecurityinsight's needs. Therefore, if another organization uses the dataset, they need to align with the label definitions provided here or relabel relevant subsets of the dataset for their needs if the definitions do not align. We hope that this disclosure will be helpful to any user of this dataset in the future.

\section*{Ethical Considerations}
The dataset is created using publicly available news articles, and does not breach any contract for obtaining the data. We have ensured that the web scraper only accesses publicly available data and excludes any data behind a paywall. Furthermore, we release only the links to the article, along with a scraper code. Therefore, if any source website changes its web scraping policies in the future, that change will be reflected when retrieving the articles. Although we cannot guarantee that PII (personally identifiable information) is not included in any of the news articles, we only source from published, publicly available materials.
For the annotation process, we leveraged internal humanitarian experts at the partnering organization \insecurityinsight, who were duly compensated for their services during the course of their professional, paid employment.

\section*{Acknowledgements}
We thank our colleagues Sirene Abou-Chakra and Jessie End from Dataminr's AI for Social Good Program for facilitating this research, Timothy Bishop, Helen Buck and domain experts from Insecurity Insight for coordinating and performing the data labeling process, and the anonymous reviewers for their constructive comments and suggestions.

\bibliography{custom}

\clearpage

\appendix
\section{Data Collection Details}
\label{sec:appendix_data}

\subsection{NewsAPI}
\label{sec:news_api}
News articles are collected from NewsAPI by combining and querying a list of keywords on a daily basis. Keyword examples are listed in Table \ref{table:newsapi_keywords}.

\begin{table}[ht]
\small{
\begin{tabular}{p{3cm}p{3cm}}
\toprule
Nouns & Verbs \\
\midrule
aid worker & abducted \\
cattle & assaulted \\
doctor & attacked \\
doctor & bombed \\
farm & burned \\
grain & damaged \\
health facility & destroyed \\
market & injured \\
mobile clinic& killed \\
NGO staff member & looted \\
refugee & robbed \\
school & shelled \\
shepherd & threatened \\
\bottomrule
\end{tabular}
}
\caption{NewsAPI query keywords for identifying violent incidents. A full list is available
\href{https://github.com/dataminr-ai/humvi-dataset/blob/main/data_collection/newsapi_keywords.json}{here}
} 
\label{table:newsapi_keywords}
\end{table}

\subsection{GDELT}
We collect all articles between December 1, 2023 and April 15, 2024 from the GDELT Events Database (version 2) using the GDELT Python library. We identify the countries associated with an article considering the following GDELT country codes: \textit{Actor1CountryCode, Actor2CountryCode, Actor1Geo\_CountryCode, Actor2Geo\_CountryCode, ActionGeo\_CountryCode}. An article's language is identified by scraping the article content and applying the ELD Python library. The complete GDELT data collection script is shared \href{https://github.com/dataminr-ai/humvi-dataset/blob/main/data_collection/gdelt_dataset.py}{here}.

\subsection{Unlabeled Dataset}
Along with \textsc{HumVI}\xspace-\expansion, we also release a large collection of unlabeled data. We believe that unlabeled data can be used to improve the model's performance through active learning~\cite{settles2009active} or through semi-supervised learning~\cite{learning2006semi}.

\section{Annotation Guidance}
\label{sec:appendix_annotation}
All annotators who participated in both the on the job labeling (\textbf{OTJ}) and offline labeling (\textbf{OFL}) were hired and trained directly by \insecurityinsight.
Tables \ref{table:labels_extended}, \ref{table:labels_extended_2}, and \ref{table:labels_extended_3} show the extended category descriptions that are used to train the annotators to perform the labeling task.

\begin{table*}
\centering
\small{
\begin{tabular}{lp{13cm}}
\toprule
Category & Description \\
\midrule
Aid Security & Harms or threats towards aid agencies or old workers. Such as:

\begin{itemize}
    \item An aid worker or staff member being killed, wounded, kidnapped, arrested, threatened, sexually harassed, discriminated, tortured, expelled, robbed or denied passage;
    \item  and/or property, vehicle, cash, equipment, or supplies belonging to an aid agency is attacked, denied access, set on fire, taken over, invaded, broken into, affected by a demonstration, have weapons installed in it, used to launch attacks, robbed of cash, equipment, or information
    \item and/or when aid agencies are forced to close or temporarily suspend operations, denied access, accused and/or investigated and/or fined, faced with demonstrations or rioting, denied a visa, deprived of utilities, faced with diversions of aid, have staff expelled, operate in a context of active fighting instability, affected by new laws or indirect government action, affected by mining, affected by strikes, be threatened;

\end{itemize}

which may result in the aid agency to proactively or reactively increase vigilance, change security measures, restrict movement, relocate staff, move assets, close, hibernate, or operate remotely.

\\
\midrule

Education & Harms or threats towards education providers, education workers, education-related infrastructure, and students. Such as:

\begin{itemize}
    \item An educator was killed, wounded, kidnapped, arrested, threatened, sexually harassed, discriminated, tortured, expelled, robbed or denied passage;
    \item and/or an education facility (school, university, any building serving to provide education services) is attacked, Set on fire, taken over, invaded, broken into, affected by a demonstration, have weapons installed in it, used to launch attacks; robbed of cash, equipment, or information
    \item or when an education provider is forced to close or temporarily suspend operations, denied access, accused and/or investigated and/or fined, faced with demonstrations or rioting, denied a visa, deprived of utilities, faced with diversions of aid, have staff expelled, operate in a context of active fighting instability, affected by new laws or indirect government action, affected by mining, affected by strikes, threatened
\end{itemize}

which may result in the education provider to proactively or reactively increase vigilance, change security measures, restrict movement, relocate staff, move assets, close, hibernate, or operate remotely.

\\
\\

& The following events are \textit{excluded}:
\begin{itemize}
    \item Events affecting retired educators, including teachers, academics, education support and transport staff.
    \item Threats or violence towards students in public/open spaces where it is clear they were not en route to or from school. E.g. kidnapped while at a market on a Saturday.
    \item Demonstrations involving students in open spaces, such as town halls and streets.
\end{itemize}

\\
\bottomrule
\end{tabular}
}
\caption{Extended description of categories}
\label{table:labels_extended}
\end{table*}

\begin{table*}
\centering
\small{
\begin{tabular}{lp{13cm}}
\toprule
Category & Description \\
\midrule

Health &  Harm or threats towards health providers or health workers. Such as:

\begin{itemize}
    \item  A health worker (doctor, nurse, ambulance driver, paramedic, military medic, physiotherapist, vaccination worker, or any other health staff member) was killed, wounded, kidnapped, arrested, threatened, sexually harassed, discriminated, tortured, expelled, robbed or denied passage;
    \item and/or Hospital, health centre, mobile health unit, pharmacy, ambulance, or other building/vehicle used for health purposes, or cash, equipment, or supplies belonging to a health provider is, attacked, denied access, set on fire, taken over, invaded, broken into, affected by a demonstration, have weapons installed in it, used to launch attacks; robbed of cash, equipment, or information
    \item or when a health provider is forced to close or temporarily suspend operations, denied access, accused and/or investigated and/or fined, faced with demonstrations or rioting, denied a visa, deprived of utilities, faced with diversions of aid, have staff expelled, operate in a context of active fighting instability, is affected by new laws or indirect government action, is affected by mining, is affected by strikes, is threatened
\end{itemize}

which may result in the the health provider to proactively or reactively increase vigilance, change security measures, restrict movement, relocate staff, move assets, close, hibernate, or operate remotely.

\\

\midrule
Protection & Harm or threats towards internally displaced persons and refugees. 

\begin{itemize}
    \item Incidents where camp settlements (both informal and formal) and its infrastructure (incl. Education and health facilities inside the camp) are affected and/or residents inside these camps are affected.
    \begin{itemize}
        \item An IDP/displaced/refugee camp was attacked, denied access, set on fire, taken over, invaded, broken into, affected by a demonstration, have weapons installed in it or armed entry into the camps, used to launch attacks, robbed of cash, equipment, or information/ looted, raided by security/military forces, forced to close or be dismantled, affected by close proximity to conflict/instability
        \item An IDP/displaced/refugee camp resident was killed, wounded, kidnapped, arrested, threatened, sexually harassed, discriminated, tortured, expelled, robbed or denied passage; forced out of camps
    \end{itemize}
    \item Events which include a disruption of aid (incl. supplies/food/security) to the camp and its residents. This includes obstruction of aid; the forced closure or dismantling of camps; inability for aid organizations to provide aid to the camps etc.
    \begin{itemize}
        \item and/or when aid agencies: are forced to close or temporarily suspend operations, denied access, accused and/or investigated and/or fined
faced with demonstrations or rioting, denied a visa, deprived of utilities, faced with diversions of aid, have staff expelled, operate in a context of active fighting instability, are affected by new laws or indirect government action, are affected by mining, are affected by strikes, are threatened.
    \end{itemize}
\end{itemize}

The following events are \textit{excluded}: 
\begin{itemize}
    \item Events where IDP/refugees/displaced persons are affected outside of the camp structure (e.g. shipwrecks; deaths on the migration routes).
    \item administrative and government actions which are aimed towards curbing migration (e.g. mass deportations; changes in government policies etc..)
\end{itemize}

\\

\bottomrule
\end{tabular}
}
\caption{Extended description of categories (continued)}
\label{table:labels_extended_2}
\end{table*}

\begin{table*}
\centering
\small{
\begin{tabular}{lp{13cm}}
\toprule
Category & Description \\
\midrule

Food Security & Harm or threats towards objects and actors directly involved in the production, processing and distribution of food; damaging or destroying transport and energy infrastructure which facilitates the food supply and distribution.

\begin{itemize}
    \item Includes events affecting objects or actors directly involved in the production, processing and distribution of food:
    \begin{itemize}
        \item The setting on fire, striking with explosive and other weapons (e.g. firearms) and damage or destruction resulting from other conflict actions to crops and grazing land, farms, food or livestock markets, food production or processing factories, food storage warehouses, granaries, farm buildings (e.g. barns and wheat storage silos, bakeries and other food shops, tractors and combine harvesters, water infrastructure (e.g. pumps, wells, pipelines and transport tanks)
        \item Looting or robbery of crops, food (including food aid), tractors, combine harvesters, agricultural equipment (e.g. ploughs, irrigators) or livestock
        \item Abductions or kidnappings, arrests and killings or injuries of: farmers and other farm workers, pastoralists or herders, fishers, food aid workers, food production workers (e.g. individuals who work in food production factories) or food distribution workers (e.g. drivers of food transport trucks)
        \item Physically violent clashes between farmers and pastoralists (including events which are both lethal and non-lethal). 
        \item Physical violence occurring at food or livestock markets (e.g. shootings using firearms, detonations of IEDs) which does not necessarily result in damage or destruction of the market itself. 
        \item Taking over, forced entry into, denial of access or invasion of farms and farm buildings, food production or processing factories, bakeries and other food shops; or food storage warehouses
    \end{itemize}
    \item Includes events which affect food security in a more indirect way by damaging or destroying transport and energy infrastructure which facilitates the continuation of food supplies and distributions.
    \begin{itemize}
        \item The setting on fire, striking with explosive weapons and conflict actions resulting in the damage or destruction of bridges, roads, ports, airports, rail lines; gas, oil and electricity pipelines and plants; or stations
    \end{itemize}
\end{itemize}

The following events are \textit{excluded}: 
\begin{itemize}
    \item protests (violent and non-violent) voicing concerns about food insecurity; non-violent political or public policies pursued by governments in relation to food insecurity (e.g. sanctions imposed on individuals, governments or private corporations); attacks on civilians except for those involved in the production of food or its distribution (e.g. farmers, pastoralists, food production factory workers).
\end{itemize}

\\

\bottomrule
\end{tabular}
}
\caption{Extended description of categories (continued)}
\label{table:labels_extended_3}
\end{table*}

\section{Modeling Details}
\subsection{Model Hyperparameters}
\label{sec:appendix_hyperparams}

For BERT, DistilBERT and RoBERTa  we leverage the transformers library and use the same training setup: the text is preprocessed with max 512 tokens. For training, we use the AdamW optimizer with $1\cdot10^{-5}$ learning rate, 16 batch size, $10$ epochs,
and 0.01 weight decay. All models were trained with the \texttt{ transformers} library~\cite{wolf2019huggingface} on the \texttt{NVIDIA A10G} machine. Each run lasted 30 minutes to 3 hours. Multiple training runs were conducted, and we report the average performance. We detail the size of each of the models in Table~\ref{table:appendix_params}.
We use the following checkpoints: 
\begin{itemize}
    \item \textsc{HumVI}\xspace-\core: bert-base-cased, distilbert-base-cased and roberta-base.
    \item \textsc{HumVI}\xspace-\expansion: bert-base-multilingual-cased, distilbert-base-multilingual-cased and xlm-roberta-base.
\end{itemize}

\subsection{LLM Prompting}
\label{sec:appendix_llm}
The prompt used in all LLM experiments is available \href{https://github.com/dataminr-ai/humvi-dataset/blob/main/llm_prompt/prompts.yaml}{here}. We experimented with multiple different prompts, and selected the final prompt due to its higher performance on a small subset and its comprehensiveness in capturing the guidelines given by \insecurityinsight to their experts. 

The size of the LLMs could be variable, and is directly related to their performance. We provide number of parameters for each open model in Table~\ref{table:appendix_params}.

\textbf{Prompt Limitations}
One of the drawbacks in our experimentation is the prompt used for LLMs. We note that in aligning the LLM prompt with the guidelines set by \insecurityinsight for their human annotators, we might have overburdened the model with context that might be distracting and might have led to sub-par zero-shot performance. We believe distilling the prompt to a smaller length, while maintaining essential guidelines could help in improving the performance. A possible direction here is to summarize the prompt, and then use it. Another possible direction is to provide examples in the prompt and leverage use few-shot learning instead of zero-shot. 

\newpage

\section{Error Analysis}

\label{sec:appendix_error}
\subsection{Misclassification Themes}
We identified two main classes of inputs for which most of the models did not perform well. 

\textbf{Missing Context}. In some cases, the \emph{victim} entity is directly related to the humanitarian aid response category and it is hard for models to understand the context. For instance, in Example~\ref{box:appendix_error1}
, the victims are employees of World Central Kitchen, an organization that provides meals to those affected by conflict. However, the models categorized them merely as aid workers, overlooking the crucial context that targeting workers of a meal-providing organization directly impacts food security.

\textbf{Indirect Impact}. For certain examples, the model focuses on the first category and ignores the rest of the text that could be detrimental for predicting other different categories that could be added for direct or indirect impact. An example is shown in Example~\ref{box:appendix_error2}. In the example, since it was an attack on UNICEF supplies, the model was able to identify the context of aid workers or their aid being targeted, but not the indirect impact of that aid being blocked which is lack of crucial health supplies and food supplies.

\subsubsection{Example of Missing Context Misclassification}
\begin{minipage}{0.9\columnwidth}
\begin{center}
\vspace{5pt}
\begin{tcolorbox}[width=\columnwidth]
\textbf{News Article}: US Secretary of State Antony Blinken said on Tuesday that Washington has urged Israel to conduct a swift, thorough and impartial investigation into Monday night’s air strike that killed seven aid workers with the World Central Kitchen charity in Gaza.

..... he said of the NGO workers killed in the strike. 
....
over the accidental deaths of seven World Central Kitchen (WCK) employees in Monday’s strike.
.... take immediate steps to protect aid workers and facilitate vital humanitarian operations in Gaza....   \\
\textbf{Ground Truth Label}:  food-sec, aid-sec\\
\textbf{Predicted Label}: aid-sec\\
\end{tcolorbox}
\vspace{5pt}
\end{center}
\label{box:appendix_error1}
\end{minipage}

\begin{table}
\centering
\small{
\begin{tabular}{cc}
\toprule
Architecture & Num Params \\
\midrule
\texttt{BERT} & 110M \\
\texttt{DistilBERT} & 66M \\
\texttt{RoBERTa} & 110M \\
\texttt{XLM-RoBERTa} & 125M \\
\texttt{DBRX} & 132B \\
\texttt{llama-3-70B} & 70B \\
\texttt{Mistral-Large} & Proprietary \\
\texttt{GPT-4-Turbo} & Proprietary \\
\texttt{GPT-4o} & Proprietary \\
\bottomrule
\end{tabular}
}
\caption{Model Size (by number of parameters) of the models we used in experiments.}
\label{table:appendix_params}
\end{table}

\subsubsection{Example of Indirect Impact Misclassification}
\begin{minipage}{0.9\columnwidth}
\begin{center}
\vspace{5pt}
\begin{tcolorbox}[width=\columnwidth]
\textbf{News Article}: UNICEF said one of their containers carrying essential supplies was looted by gangs at Haiti's main port. ....
The United Nations Children's Fund (UNICEF) said Saturday that one of its 17 aid containers at Haiti's main port was looted.
....
The container was carrying "essential items for maternal, neonatal, and child survival, as well as critical supplies for early childhood development and education, water equipment, and others," the agency said.
.....
"Looting of supplies that are essential for life saving support for children must end immediately," ....
Gang violence has spiked throughout the country in recent days.
....
Some hospitals in the city have been forced to close over safety concerns....
Shortages of electricity, fuel and medical supplies have affected hospitals ....
is supported by the Caribbean regional body CARICOM, the United Nations and the United States.   \\
\textbf{Ground Truth Label}:  food-sec, aid-sec, health\\
\textbf{Predicted Label}: aid-sec\\
\end{tcolorbox}
\vspace{5pt}
\end{center}
\label{box:appendix_error2}
\end{minipage}

\end{document}